\documentclass[runningheads]{llncs}

\usepackage[T1]{fontenc}
\usepackage{graphicx}
\usepackage{booktabs}
\usepackage{multirow}
\usepackage{amsmath}
\usepackage{microtype}
\usepackage{xcolor}
\usepackage[hidelinks]{hyperref}

\makeatletter
\renewcommand\subsection{\@startsection{subsection}{2}{\z@}%
                       {-11\p@ \@plus -3\p@ \@minus -3\p@}%
                       {4\p@ \@plus 2\p@ \@minus 2\p@}%
                       {\normalfont\normalsize\bfseries\boldmath
                        \rightskip=\z@ \@plus 8em\pretolerance=10000 }}
\renewcommand\subsubsection{\@startsection{subsubsection}{3}{\z@}%
                       {-10\p@ \@plus -3\p@ \@minus -3\p@}%
                       {-0.4em \@plus -0.18em \@minus -0.08em}%
                       {\normalfont\normalsize\bfseries\boldmath}}
\renewcommand\paragraph{\@startsection{paragraph}{4}{\z@}%
                       {-7\p@ \@plus -3\p@ \@minus -3\p@}%
                       {-0.4em \@plus -0.18em \@minus -0.08em}%
                       {\normalfont\normalsize\itshape}}
\makeatother

\newcommand{\method}{PriorHybrid-CBM}

\begin{document}

\title{Less Annotation, More Interpretation: Prior-Guided Concept Bottleneck Models for Interpretable Cancer Imaging Diagnosis}

\titlerunning{Prior-Guided CBMs for Cancer Imaging}

\author{Baoqiang Ma\inst{1}\thanks{Corresponding author.} \and
Kenneth Gilhuijs\inst{1}}
\authorrunning{B. Ma and K. Gilhuijs}
\institute{Image Sciences Institute, University Medical Center Utrecht,
Utrecht, The Netherlands\\
\email{B.Ma-2@umcutrecht.nl}}

\maketitle

\begin{abstract}
Concept bottleneck models (CBMs) can improve the transparency of cancer image diagnostic prediction by expressing predictions through radiological concepts. However, their dependence on instance-level concept annotations limits practical applicability. We propose a prior-guided hybrid CBM that integrates limited concept annotations, class-conditional concept distribution matching on unannotated patients, and prior initialization of the concept-to-diagnosis head. We evaluate the method on CBIS-DDSM mammographic masses and calcifications and LIDC-IDRI pulmonary nodules across 0--100\% concept annotation. In the clinically relevant 0--20\% annotation regime, the hybrid CBM consistently improves mean concept AUC over a matched standard CBM, while maintaining diagnostic performance close to black-box models. At 10\% annotation specifically, concept AUC increases from 0.619 to 0.741 for masses, from 0.650 to 0.787 for calcifications, and from 0.597 to 0.642 for pulmonary nodules. Ablation experiments identify prior initialization as the main component contributing to improved concept detection, likely by stabilizing the concept-to-diagnosis head. Zero-shot VLMs remain insufficient for reliable fine-grained tumor-level concept prediction. These findings suggest that structured priors can substantially reduce the annotation burden of interpretable cancer imaging models.


\end{abstract}

\section{Introduction}

Radiology imaging-based deep learning models have demonstrated strong
performance in supporting cancer diagnosis and treatment decision-making
\cite{ardila2019lung,mckinney2020breast}. However, their clinical adoption
remains limited due to the lack of relevant evidence and the unclear
decision-making process underlying their predictions. Therefore, many
explainable artificial intelligence (XAI) methods have been proposed to
investigate their interpretability. Saliency-based XAI methods, such as
Grad-CAM \cite{selvaraju2017gradcam} and Integrated Gradients
\cite{sundararajan2017integrated}, have been widely applied to localize image
regions contributing to model predictions. The choice of XAI methods also substantially affect explanation quality in medical imaging \cite{ma2026rankingxai}. However, they primarily indicate
where a model focuses rather than what radiological
characteristics inform its decisions.

Concept bottleneck models (CBMs) offer intrinsic interpretability by first
predicting human-understandable attributes from input images and subsequently
using these attributes for final outcome prediction \cite{koh2020cbm}. In
cancer imaging, these attributes may correspond to radiological
characteristics such as spiculation, lesion shape, margin characteristics,
and calcification morphology. CBMs then assign different weights to these
attributes when generating the final diagnosis. This mimics the
radiological diagnostic workflow. However, a major limitation of CBMs is
their reliance on image-level concept labels for training, which require
expensive and time-consuming expert annotations.

Several approached have been proposed to reduce this burden. For example,
semi-supervised CBMs leverage additional unlabeled images through
pseudo-labeling or concept-level alignment to improve concept learning
\cite{hu2024sscbm}. Another direction exploits prior clinical knowledge to
guide concept prediction. Instead of using image-level concept annotations,
Nahiduzzaman et al. utilized predefined class-level concept priors to
weakly supervise concept learning \cite{nahiduzzaman2025priors}. Recently,
vision-language models (VLMs) have emerged as a promising alternative for
label-efficient concept learning. By learning aligned image and text
representations, VLMs enable zero-shot (e.g., label-free CBMs
\cite{oikarinen2023lfcbm}) or few-shot (e.g., CBVLM
\cite{patricio2025cbvlm} for lung nodule diagnosis \cite{ma2026vlmlung}) concept prediction without extensive concept
annotations. In particular, CLIP-like models have been widely adopted for
zero-shot concept detection by measuring the similarity between image and
concept text representations \cite{radford2021clip}. Several CLIP-based models
have been developed for radiological imaging. For example, Mammo-CLIP learns
joint representations from mammogram-report pairs
\cite{ghosh2024mammoclip} and CT-CLIP aligns 3D chest CT volumes with
radiology reports \cite{hamamci2025ctclip}. These models demonstrate that
broad clinically meaningful information, often at the organ level, can be
acquired directly from medical image-report pairs. However, their
effectiveness in detecting fine-grained tumor-level radiological concepts,
such as lesion morphology and margin characteristics remains insufficiently
explored.

To address this gap, we investigate concept learning for fine-grained
tumor-level radiological concepts under varying levels of supervision.
Specifically, we:
\begin{enumerate}
    \item Conduct a systematic comparison of concept learning under partial
    supervision, prior-guided learning, and zero-shot VLM transfer on the
    CBIS-DDSM \cite{lee2017cbis} and LIDC-IDRI \cite{armato2011lidc} datasets;
    \item Propose a hybrid CBM that bridges partial concept supervision and
    prior-guided learning, substantially improving concept prediction in
    low-annotation settings;
    \item Demonstrate that clinically informed prior initialization stabilizes
    the concept-to-diagnosis mapping and enhances concept learning, especially
    when concept annotations are scarce;
    \item Show that current modality-specific VLMs remain insufficient for
    reliable zero-shot prediction of subtle tumor-level radiological concepts.
\end{enumerate}

\section{Method}

\begin{figure}[t]
\centering
\includegraphics[width=\textwidth]{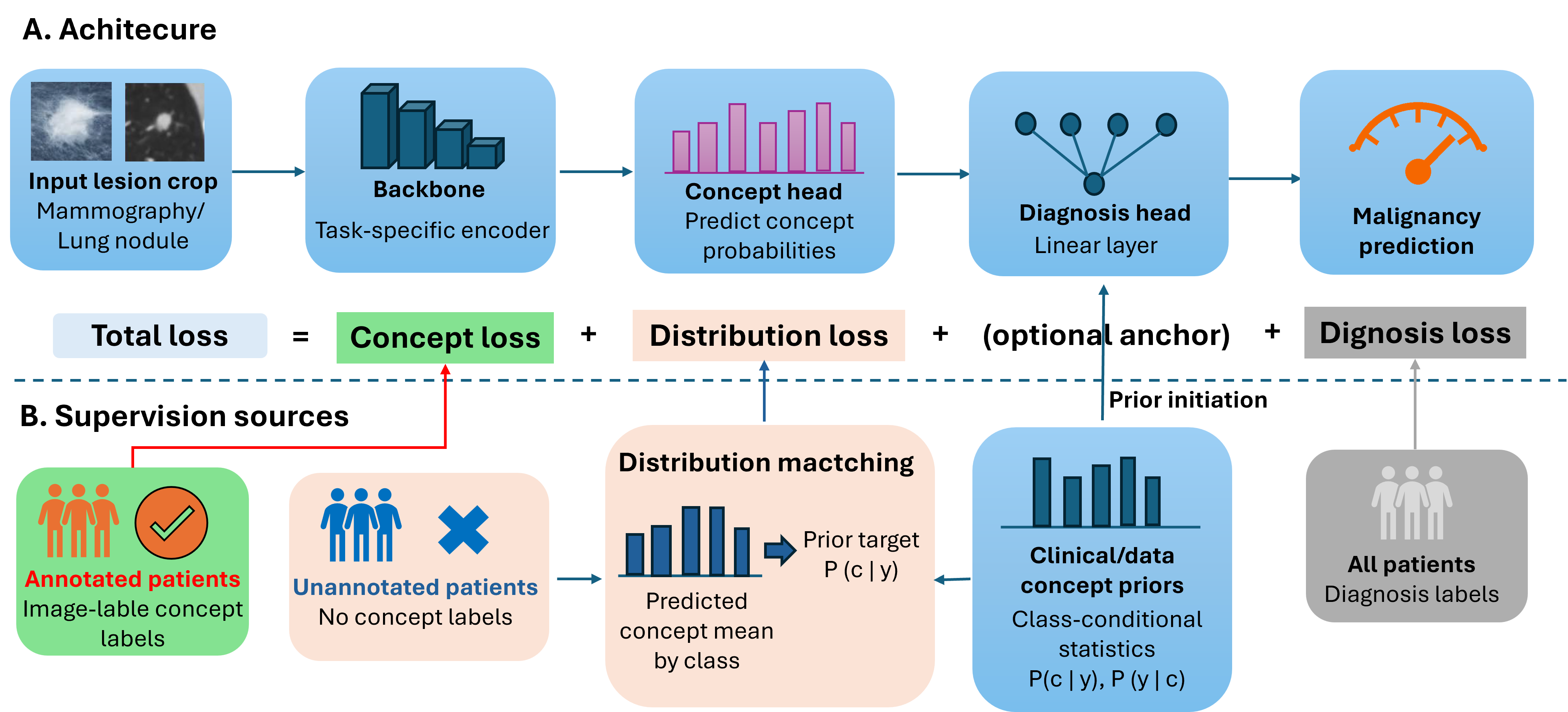}
\caption{Overview of \method. A task-specific image backbone predicts
radiological concepts, which are used by a linear diagnosis head. Patients with
concept annotations receive instance-level concept supervision; patients
without concept annotations receive class-conditional prior supervision.
Prior concept knowledge also initializes (with optional anchors) the
concept-to-diagnosis head.}
\label{fig:framework}
\end{figure}

\subsection{Overview of Hybrid Concept Bottleneck Model}

The proposed prior-guided hybrid CBM is constructed based on the standard CBM
architecture. As shown in Fig.~\ref{fig:framework}A, given an input image $x$, a
task-specific backbone is applied to extract image features, followed by a linear concept head that produces logits $\mathbf z=g_\theta(x)$ and
probabilities of predefined concepts,
$\hat{\mathbf c}=\sigma(\mathbf z)\in[0,1]^K$. A linear diagnosis head then
predicts malignancy probability using these concept probabilities.

When concepts annotations are available for all patients, a standard CBM is trained using diagnosis and
concept losses:
\begin{equation}
\mathcal L_{\mathrm{CBM}} =
\mathcal L_{\mathrm{diag}}(\mathcal D)
+\lambda_c\mathcal L_{\mathrm{concept}}(\mathcal D),
\label{eq:standard-cbm}
\end{equation}
where $\mathcal D$ denotes all training samples. This objective is
interpretable but annotation-expensive in clinical imaging.

To reduce the concept annotation cost and keep interpretability, we train the CBM in a hybrid way. Let
$\mathcal A$ denote patients with instance-level concept annotations and
$\mathcal U$ the remaining unannotated patients. For annotated patients, we
use the concept loss. For unannotated
patients, we use concept distribution matching loss to transfer class-conditional
prior knowledge as weak concept supervision. In addition, the same prior
knowledge is used to initialize the weights of the diagnosis head with clinical meaning. And we can optionally regularize these weights toward this clinically meaningful initialization. Therefore, the resulting
hybrid objective is
\begin{equation}
\mathcal L =
\mathcal L_{\mathrm{diag}}(\mathcal D)
+\lambda_c\mathcal L_{\mathrm{concept}}(\mathcal A)
+\lambda_d\mathcal L_{\mathrm{dist}}(\mathcal U)
+\lambda_r\mathcal L_{\mathrm{reg}}.
\label{eq:objective}
\end{equation}
$\mathcal L_{\mathrm{diag}}$ is the weighted binary cross-entropy for malignancy
in all samples. $\mathcal L_{\mathrm{concept}}$ is binary cross-entropy with
logits between $\mathbf z_i$ and the image-level concept labels
$\mathbf c_i$, evaluated only for samples in $\mathcal A$.  $\mathcal L_{\mathrm{dist}}$ is illustrated in Section 2.2. The
standard baseline CBM can be considered a special setting of this hybrid CBM:
setting $\lambda_d=\lambda_r=0$ and using random initialization
for the diagnosis head.

\subsection{Class-Conditional Prior Supervision}

When concept labels are missing, the model can still be guided by the frequency
with which radiological concepts are expected to occur in benign and malignant cases.
This converts clinical or cohort-level knowledge into a weak supervision
signal for the concept predictor training. This relates to learning from aggregate
label proportions \cite{quadrianto2009label} rather than treating
unannotated patients as uninformative.

For diagnosis class $y\in\{0,1\}$, a prior vector
$\boldsymbol\pi_y\in[0,1]^K$ encodes the expected class-conditional concept
prevalence $P(c_k=1\mid y)$. For the unannotated samples of class $y$ within a
minibatch, distribution matching loss compares the mean predicted concept vector
with this prior:
\begin{equation}
\mathcal L_{\mathrm{dist}} =
\frac{1}{|\mathcal Y_B|}
\sum_{y\in\mathcal Y_B}
\mathrm{BCE}
\left(
\frac{1}{|\mathcal U_y|}\sum_{i\in\mathcal U_y}\hat{\mathbf c}_i,
\boldsymbol\pi_y
\right),
\label{eq:dist}
\end{equation}
where $\mathcal Y_B$ contains the classes present among the unannotated
samples in the minibatch. The loss is applied only to $\mathcal U$ unannotated patients.  This
aggregate prior is used to fill the supervision gap for patients without
concept annotations.

We evaluate two prior sources: (1) Manual clinical priors encode broad
BI-RADS- or radiology-based associations but are not calibrated to the specific
dataset population. In contrast, (2) data priors estimate $P(c_k=1\mid y)$ from training 
patients and therefore better match the class prevalence of concepts of the specifc dataset. 

\subsection{Prior Initialization and Optional Anchoring}

In addition to using prior knowledge for weakly concept supervision, we use $P(y\mid c)$ to initialize
the diagnosis head. This gives the model a clinically plausible concept-to-diagnosis mapping. For concept $k$, the initial weight is
\begin{equation}
w_k^{(0)} =
\mathrm{logit}\{P(y=1\mid c_k=1)\}
-
\mathrm{logit}\{P(y=1\mid c_k=0)\},
\label{eq:init}
\end{equation}
and the bias is initialized to the logit of the malignancy base rate. We also
evaluate an optional anchor toward these initial weights,
\begin{equation}
\mathcal L_{\mathrm{reg}}=
\frac{1}{K}\|\mathbf w-\mathbf w_{\mathrm{prior}}\|_2^2 .
\label{eq:reg}
\end{equation}
In the main configuration, we use a weak anchor with $\lambda_r=0.05$ and
evaluate the no-anchor setting ($\lambda_r=0$) as a component ablation.

\section{Experiments}

\subsection{Datasets and Preprocessing}

CBIS-DDSM \cite{lee2017cbis} consists of digitized mammography images with annotated diagnosis labels and BI-RADS-related lesion descriptors. The mass cohort of it comprised 1,318 training (regions of interest) ROIs images from 691 patients and 377 held-out test ROIs from 201 patients and concepts describing lesion shape, margin and breast density. The calcification cohort comprised 1,545 training ROIs from 602 patients and 324 held-out test ROIs from 151 patients and concepts describing morphology, distribution, and breast density. ROI images were min--max normalized, padded to a square field of view, resized to $224\times224$, and loaded as three-channel images with ImageNet normalization. Diagnosis labels were binarized as malignant versus benign or benign-without-callback. All lesion descriptors were transformed into 18 or 14 multi-hot encoding for mass or calcification cases, with the exception of breast density being mapped to $[0,1]$. 

Pulmonary nodule malignancy was evaluated on LIDC-IDRI CT scans \cite{armato2011lidc}. Nodules with median malignancy score 3 were excluded, and malignancy was binarized as median score $>3$ versus $<3$. The remaining cohort was split at the patient level into a training set of 1,204 nodules from 589 patients and a held-out test set of 213 nodules from 105 patients. Nodule-centered volumes of $48^3$ were cropped from pre-resampled 1mm CT, clipped to $[-1000,400]$ HU, and scaled to $[0,1]$.  Seven binary radiological concepts were derived by thresholding median radiologist scores or size: obvious nodule, calcification, round shape, well-defined margin, lobulation, spiculation, and large size. These corresponded to subtlety $\geq3$, calcification $<6$, sphericity $\geq4$, margin $\geq4$, lobulation $\geq3$, spiculation $\geq3$, and diameter $\geq10$ mm.

For all three tasks, training patients were partitioned into five folds for model validation and selection.

\subsection{Implementation and Evaluation}

CBIS-DDSM models employed an ImageNet-pretrained 2D DenseNet121 \cite{huang2017densely} as backbone architecture applied to mammographic ROIs, whereas LIDC-IDRI models used a Med3D-pretrained 3D ResNet18 \cite{chen2019med3d} applied to nodule crops. The concept bottleneck model then include a dropout layer, a linear concept head with sigmoid activations, and a linear diagnosis head operating on the predicted concept probabilities. All models were optimized with AdamW using learning rate $10^{-4}$, weight decay $10^{-5}$, dropout 0.5, batch size 32, and up to 100 epochs. The learning rate was reduced on validation AUC plateaus. Checkpoints were selected by validation diagnostic AUC after a 30-epoch warm-up, with early stopping patience of 20 epochs. Data augmentation was applied. For the main prior-guided configuration, we swept $\lambda_d\in\{0,0.3,0.7,1.0\}$ with data-prior initialization and data-derived distribution targets. The task-specific $\lambda_d$ was selected by mean five-fold validation concept AUC over the 0--20\% annotation regime after the warm-up period, giving $\lambda_d=0.7$ for CBIS-DDSM mass, $\lambda_d=0.3$ for CBIS-DDSM calcification, and $\lambda_d=1.0$ for LIDC-IDRI. For each concept-annotation fraction, five fold models were trained and evaluated on the same held-out test set. Final test predictions were obtained by averaging the five models' malignancy and concept probabilities. Concept annotations were randomly sampled at the patient level at fractions of $0,5,10,20,50,75$, and $100\%$. These fractions applied only to labels used in the instance-level concept loss; data-derived aggregate priors were computed separately from the training patients of each fold. The implementation code is available at \url{https://github.com/baoqiangma96/prior-guided-hybrid-cbm}. 

The primary evaluation metrics were diagnostic ROC-AUC and mean per-concept ROC-AUC. 

\subsection{Compared Methods}
We compare the proposed prior-guided hybrid CBM with three references. The
black-box model uses the same dataset-specific backbone followed by a linear
diagnosis head. The standard CBM uses the same
CBM architecture as the hybrid model,
but removes prior guidance by using random diagnosis-head initialization and
setting $\lambda_d=\lambda_r=0$. Zero-shot VLM references use Mammo-CLIP
\cite{ghosh2024mammoclip} for CBIS-DDSM and CT-CLIP
\cite{hamamci2025ctclip} for LIDC-IDRI to predict concept probabilities from
positive/negative text prompts; these concept probabilities are then passed
through frozen standard CBM diagnosis heads for malignancy prediction. 

\subsection{Ablation Study}

The ablation study focuses on 0, 5, 10,
and 20\% concept annotation and uses the same task-specific selected
distribution weights.

Starting from the main prior-guided configuration, which uses data-prior
initialization, data-derived distribution matching, and continued L2 anchoring
with $\lambda_r=0.05$, we perform three component-removal ablations: removing the distribution loss ($\lambda_d=0$), removing prior initialization by using random diagnosis-head
initialization, and removing the L2 anchoring loss ($\lambda_r=0$). These
ablations test the contributions of distribution-level concept supervision,
prior-informed concept-to-diagnosis initialization, and continued anchoring to
the initialized diagnosis head, respectively.

\section{Results and Discussion}

\subsection{Annotation-Efficient Concept Learning}

\begin{figure}[t]
\centering
\includegraphics[width=\textwidth]{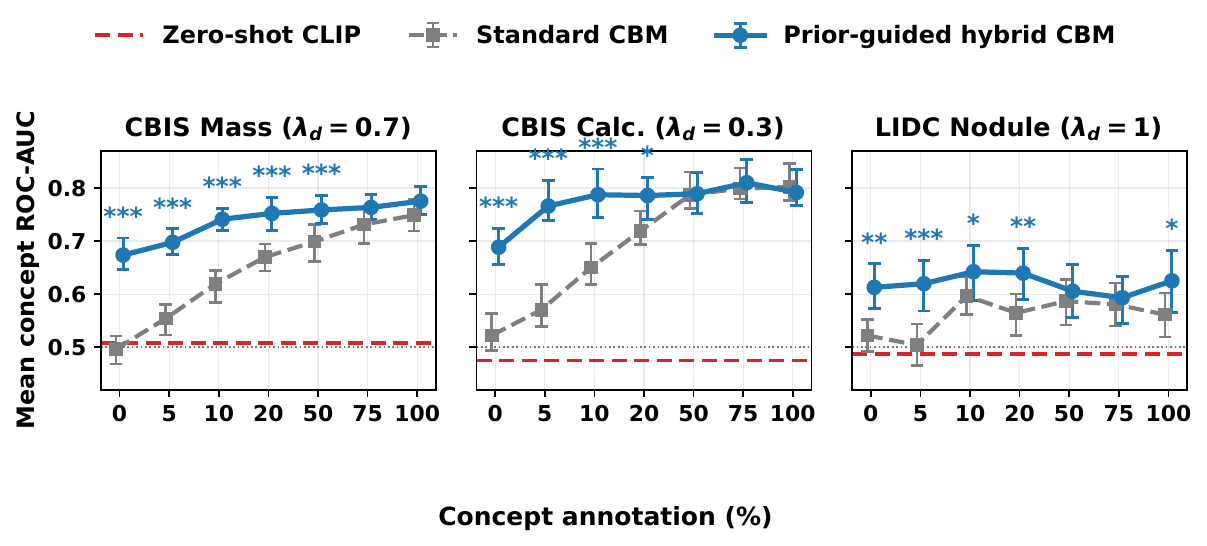}
\caption{Concept annotation efficiency. Mean concept ROC-AUC is reported for
the prior-guided hybrid CBM, matched standard CBM, and zero-shot VLM reference.
Error bars denote percentile bootstrap 95\% confidence intervals. Stars denote
paired bootstrap significance for hybrid CBM versus standard CBM.}
\label{fig:concept-efficiency}
\end{figure}

Figure~\ref{fig:concept-efficiency} shows that prior guidance mainly improves
the clinically relevant low-annotation regime. Between 0 and 20\% concept
annotation, the prior-guided hybrid CBM consistently outperformed the matched
standard CBM in mean concept ROC-AUC significantly. At 10\% annotation, concept AUC increased
from 0.619 to 0.741 for CBIS-DDSM masses, from 0.650 to 0.787 for
calcifications, and from 0.597 to 0.642 for LIDC-IDRI nodules. This is the
central practical benefit of the method: radiologist concept annotations are
expensive, and the largest gains appear exactly when only a small subset of
patients has instance-level concept labels.

The zero-shot VLM references were consistently weaker for concept prediction, achieving AUCs around 0.50. Therefore, prompt-based annotation-free concept recognition is not effective enough in these cancer-imaging tasks.

\subsection{Diagnostic Performance}

\begin{figure}[t]
\centering
\includegraphics[width=\textwidth]{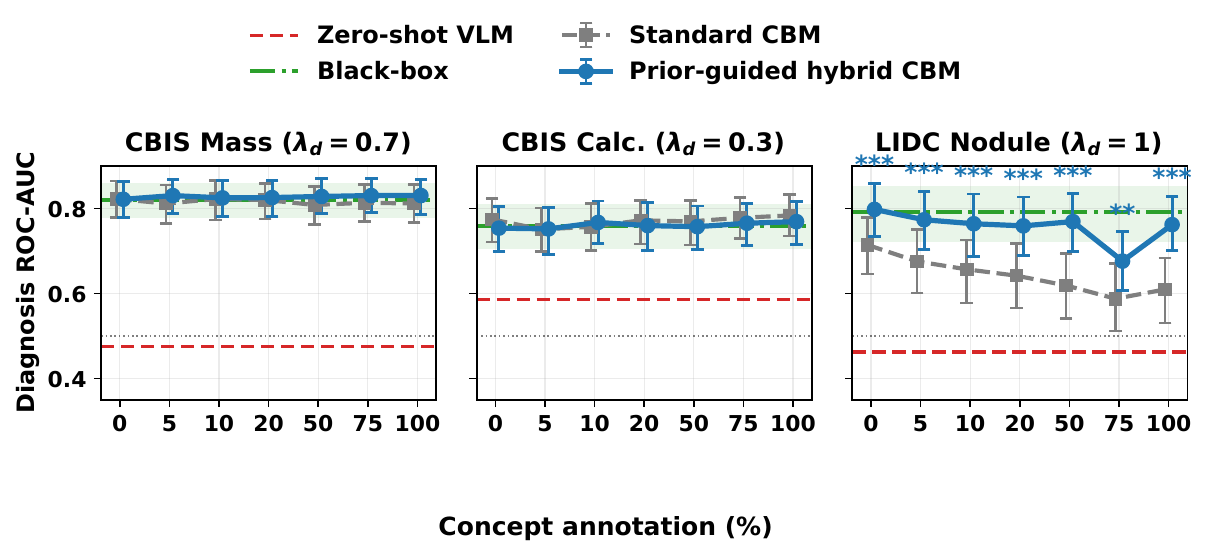}
\caption{Diagnostic annotation efficiency. Diagnosis ROC-AUC is shown for the
prior-guided hybrid CBM, matched standard CBM, black-box reference, and
zero-shot VLM reference. Error bars denote percentile bootstrap 95\%
confidence intervals for trained CBM models; the black-box reference is shown
as a horizontal line with bootstrap confidence band.}
\label{fig:diagnosis-efficiency}
\end{figure}

Figure~\ref{fig:diagnosis-efficiency} shows that improved concept detection did not require sacrificing diagnostic discrimination. For CBIS-DDSM masses and calcifications, both the standard CBM and the prior-guided hybrid CBM achieved consistently high diagnosis ROC-AUC across all annotation ratios, remaining close to the black-box references. At 10\% annotation, the hybrid CBM achieved diagnosis AUCs of 0.825 for masses and 0.767 for calcifications. For LIDC-IDRI pulmonary nodules, the hybrid CBM outperformed the standard CBM across most annotation ratios and remained close to the black-box model in the 0--20\% annotation regime, achieving an AUC of 0.764 at 10\% annotation. In contrast, the zero-shot VLM baseline showed substantially lower diagnostic performance across all three tasks, indicating that zero-shot VLM concepts are not reliable enough for fine-grained tumor-level CBMs.

\subsection{Ablation and Prior Source}

Table~\ref{tab:ablation} shows the contributions of distribution matching,
prior initialization, and continued anchoring. Averaged over 0, 5, 10, and
20\% annotation. The full prior-guided model improved mean concept AUC over the
standard CBM by 0.131 for masses, 0.142 for calcifications, and 0.082 for
nodules. Removing prior initialization caused the clearest concept drop,
especially for CBIS-DDSM, while removing the L2 anchoring term had little
effect. The distribution loss had a more task-dependent effect once data-prior
initialization was retained. It improved the averaged concept AUC for
calcifications and nodules, but was similar for masses. 

\begin{table}[t]
\centering
\caption{Component ablation averaged over 0, 5, 10, and 20\% concept
annotation. Each cell reports diagnosis AUC / mean concept AUC.}
\label{tab:ablation}
\resizebox{\linewidth}{!}{%
\begin{tabular}{lccccc}
\toprule
Task & Main & $-\mathcal{L}_{dist}$ & Random init. & $-\mathcal{L}_{reg}$ & Standard CBM \\
\midrule
CBIS-DDSM mass & .825 / .716 & .823 / .719 & .824 / .653 & .829 / .716 & .819 / .585 \\
CBIS-DDSM calcification & .758 / .757 & .762 / .749 & .762 / .671 & .760 / .761 & .763 / .615 \\
LIDC-IDRI nodule & .773 / .629 & .768 / .619 & .727 / .603 & .781 / .628 & .672 / .547 \\
\bottomrule
\end{tabular}%
}
\end{table}

The full prior-source sweep in Supplementary Fig.~S1
shows that data-derived priors generally gave stronger concept supervision
than manual priors. Using the selected $\lambda_d$, mean concept AUC over
0--20\% annotation improved from 0.681 to 0.716 for masses and from 0.726 to
0.757 for calcifications when moving from manual-prior initialization and
manual distribution targets to data-prior initialization and data-derived
targets. For LIDC-IDRI, manual and data priors were similar for concept AUC
(0.626 versus 0.629). This does not mean that BI-RADS or manual radiology
knowledge is incorrect. Instead, data-derived priors are better calibrated to
the specific prevalence, label definitions, and concept co-occurrence patterns
of the benchmark cohort.

\subsection{Why Prior Initialization Matters}

The diagnosis-head drift analysis in Supplementary Table~S1 further explains why prior initialization was important. At 0\% concept annotation, diagnosis heads initialized with either the data prior or the manual prior changed only minimally from their initial weights: the L2 drift was 0.024--0.027 for masses, 0.016--0.025 for calcifications, and 0.006--0.024 for nodules, corresponding to less than 0.5\% relative drift of the full weight vector. In contrast, randomly initialized heads showed substantially larger changes, with L2 drift ranging from 0.492 to 0.877, and moved far from the empirical data-prior direction.

These results suggest that prior initialization stabilizes the concept-to-diagnosis mapping when instance-level concept supervision is absent. By starting from a clinically meaningful concept-diagnosis association rather than a random diagnostic direction, the model can learn image-to-concept representations while keeping the diagnosis head anchored to a plausible prior. This provides a more stable optimization signal when concept annotations are scarce or unavailable.

\subsection{Concept Correction and Future Clinical Use}

We further tested an oracle concept-correction setting at 10\% annotation Table~\ref{tab:concept-intervention}). The diagnosis head was fixed, and threshold-incorrect concept predictions were replaced with ground-truth labels. This simulates an idealized clinician-in-the-loop workflow in which concept errors are corrected before diagnosis. Concept correction improved diagnosis AUC by 0.011 for masses, 0.034 for calcifications, and 0.139 for LIDC-IDRI nodules. This indicates that better concept prediction and selective correction can improve downstream diagnosis.

\begin{table}[t]
\centering
\caption{Test-time concept-correction intervention for the selected
prior-guided hybrid CBM at 10\% concept annotation. The diagnosis head is
fixed; threshold-incorrect concept predictions are replaced by ground-truth
concept labels.}
\label{tab:concept-intervention}
\resizebox{\linewidth}{!}{%
\begin{tabular}{lccc}
\toprule
Task & Predicted concepts & Corrected concepts & $\Delta$ AUC \\
\midrule
CBIS-DDSM mass & .825 & .836 & +.011 \\
CBIS-DDSM calcification & .767 & .801 & +.034 \\
LIDC-IDRI nodule & .764 & .903 & +.139 \\
\bottomrule
\end{tabular}%
}
\end{table}

\subsection{Limitations}

This study has several limitations. First, both datasets are retrospective public benchmarks, and the learned data priors may encode cohort-specific correlations. Second, data-derived priors require aggregate concept statistics, potentially obtained from historical annotations. Thus, any reduction in clinical workload is conditional on a suitable prior being available and should be confirmed in prospective annotation-time studies. Third, we report mean concept AUC and did not perform a detailed per-concept error analysis. As a result, rare concepts may remain unstable even when the mean performance improves. Fourth, the concept-correction experiment is an oracle analysis, as it assumes prior knowledge of which concept predictions are incorrect. Future work should evaluate the method on prospective external cohorts, estimate priors from limited or external annotations, and test clinician-in-the-loop workflows in which radiologists selectively review and correct clinically important concepts.


\section{Conclusion}

This study demonstrates that prior-guided hybrid CBMs can reduce the need for
instance-level concept annotations in cancer imaging while preserving clinically
meaningful concept-based reasoning. By combining limited expert concept labels
with class-conditional prior supervision and prior initialization, the proposed
model improves concept detection in low-annotation settings without sacrificing
diagnostic discrimination. The results also show that current zero-shot VLMs do
not yet provide a sufficient substitute for task-specific concept supervision.
Overall, structured prior knowledge offers a practical path toward more
annotation-efficient and interpretable cancer imaging models.

\begin{credits}
\subsubsection{\discintname}
The authors have no competing interests to declare that are relevant to the
content of this article.
\end{credits}

\bibliographystyle{splncs04}
\bibliography{references}

\clearpage
\renewcommand{\thefigure}{S\arabic{figure}}
\renewcommand{\thetable}{S\arabic{table}}
\setcounter{figure}{0}
\setcounter{table}{0}
\appendix
\section{Complete Hybrid-CBM Prior and Distribution Sweep}
\label{sec:full-curves}

\begin{figure}[!htbp]
\centering
\includegraphics[width=\textwidth]{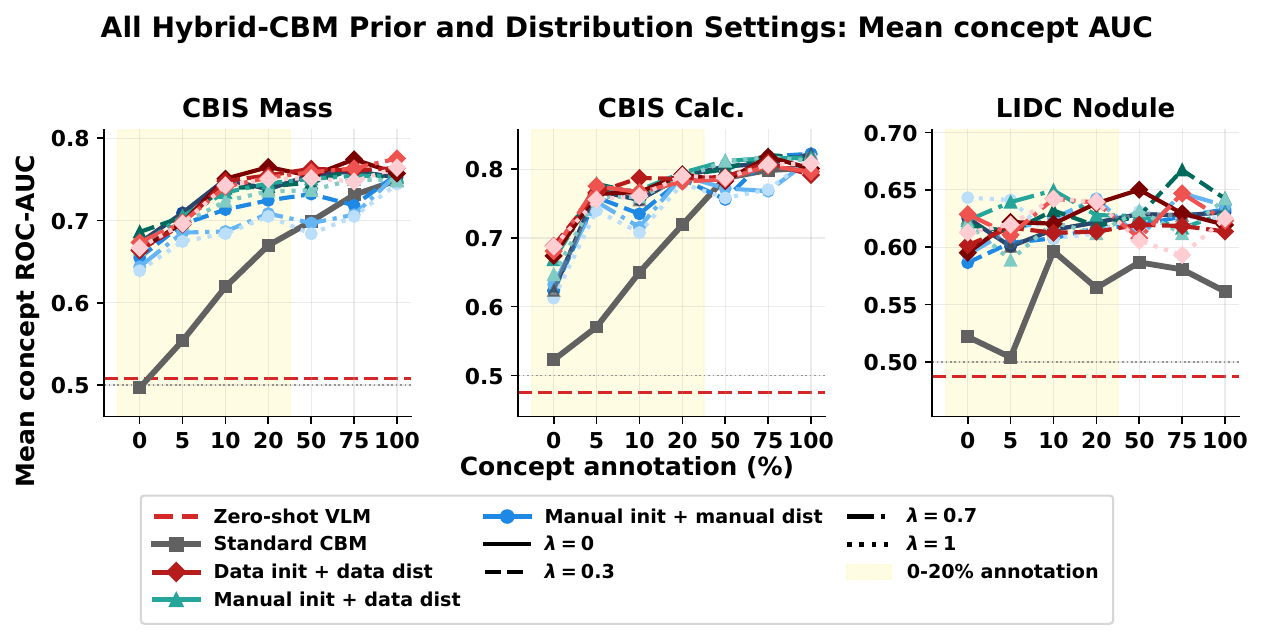}
\caption{Complete mean concept ROC-AUC sweep across evaluated prior sources
and distribution-loss weights. The shaded region denotes the 0--20\%
low-annotation regime emphasized in the main analysis.}
\label{fig:all-hybrid-concept}
\end{figure}

\section{Diagnosis-Head Weight Drift}
\label{sec:weight-drift}

\begin{table}[!htbp]
\centering
\caption{Diagnosis-head weight drift at 0\% concept annotation. Each cell
reports L2 drift from initialization, with relative drift in parentheses.}
\label{tab:weight-drift}
\resizebox{\linewidth}{!}{%
\begin{tabular}{lcccc}
\toprule
Task & Data init + anchor & Data init, no anchor & Manual init + anchor & Random init \\
\midrule
CBIS-DDSM mass & 0.024 (0.35\%) & 0.019 (0.28\%) & 0.027 (0.39\%) & 0.817 (141.51\%) \\
CBIS-DDSM calcification & 0.025 (0.22\%) & 0.024 (0.21\%) & 0.016 (0.16\%) & 0.877 (151.90\%) \\
LIDC-IDRI nodule & 0.024 (0.41\%) & 0.021 (0.36\%) & 0.006 (0.12\%) & 0.492 (85.25\%) \\
\bottomrule
\end{tabular}%
}
\end{table}

\end{document}